\documentclass[10pt, a4paper]{article}

\usepackage{lrec-coling2024} 

\usepackage{booktabs}
\usepackage{amsmath} 
\usepackage{graphicx}
\usepackage{tabularx}
\usepackage{soul}
\usepackage{natbib}
\usepackage{multibib}
\usepackage{multirow} 
\usepackage{nameref}
\usepackage{appendix}
\usepackage{xcolor,xspace}
\usepackage{enumitem}
\usepackage{bm}

\title{ChartThinker: A Contextual Chain-of-Thought Approach to Optimized Chart Summarization}

\name{Mengsha Liu$^{1}$, Daoyuan Chen$^{2}$, Yaliang Li$^{2}$, Guian Fang$^{1}$,Ying Shen$^{1}$$^{2}$$^{3}$\sthanks{\ \ Corresponding authors} }
\address{
    $^{1}$ Sun Yat-sen University, \\
    $^{2}$ Alibaba Group \\
    $^{3}$ Guangdong Provincial Key Laboratory of Fire Science \\and Intelligent Emergency Technology, Guangzhou 510006, China \\
    $^{4}$ Pazhou Lab, Guangzhou 510005, China \\
    \texttt{\{liumsh6, fanggan\}@mail2.sysu.edu.cn}\\
    \texttt{\{daoyuanchen.cdy, yaliang.li\}@@alibaba-inc.com}\\
    \texttt{sheny76@mail.sysu.edu.cn}
    }

\abstract{Data visualization serves as a critical means for presenting data and mining its valuable insights. The task of chart summarization, through natural language processing techniques, facilitates in-depth data analysis of charts. However, there still are notable deficiencies in terms of visual-language matching and reasoning ability for existing approaches. 
To address these limitations, this study constructs a large-scale dataset of comprehensive chart-caption pairs and fine-tuning instructions on each chart.
Thanks to the broad coverage of various topics and visual styles within this dataset, better matching degree can be achieved from the view of training data.
Moreover, we propose an innovative chart summarization method, ChartThinker, which synthesizes deep analysis based on chains of thought and strategies of context retrieval, aiming to improve the logical coherence and accuracy of the generated summaries. 
Built upon the curated datasets, our trained model consistently exhibits superior performance in chart summarization tasks, surpassing 8 state-of-the-art models over 7 evaluation metrics. Our dataset and codes are publicly accessible.
 \\ \newline  }

\begin{document}

\maketitleabstract

\section{Introduction}



Data visualizations, such as bar charts and line charts, are widely used to present quantitative data. These charts are valuable tools for gaining insights from data and making informed decisions. However, manually writing textual descriptions for charts can be time-consuming and prone to errors \cite{stokes2022striking}. Automatic chart summarization addresses this challenge by explaining a chart and summarizing its key takeaways in natural language. Using such systems, not only can the interpretability of the charts be enhanced, but they can also significantly reduce the time and cognitive resources required, thereby optimizing workflow efficiency \cite{obeid2020chart}.

In the early stages, researchers relied on template-based methods that combined statistical techniques and planning-based architecture to generate captions for charts \cite{reiter2007architecture}.  However, this method has its limitations, as it often leads to similar answer styles across different charts. More recently, there has been a shift towards exploring data-driven neural models for describing tabular data \cite{liu2018table}.  This approach involves converting all charts into tables and then transforming these tables into descriptive texts \cite{liu2022deplot}. Although this approach can accurately capture the data within the charts, 
it also results in the omission of a substantial amount of information, such as chart types, curve trends, and other crucial details. Furthermore, with the advancement of large visual-language models, some researchers have begun utilizing pre-trained models trained on language and vision tasks to address the chart-to-text task \cite{masry2023unichart}.

\begin{figure*}[h!]
  \centering
  \includegraphics[width=\textwidth]{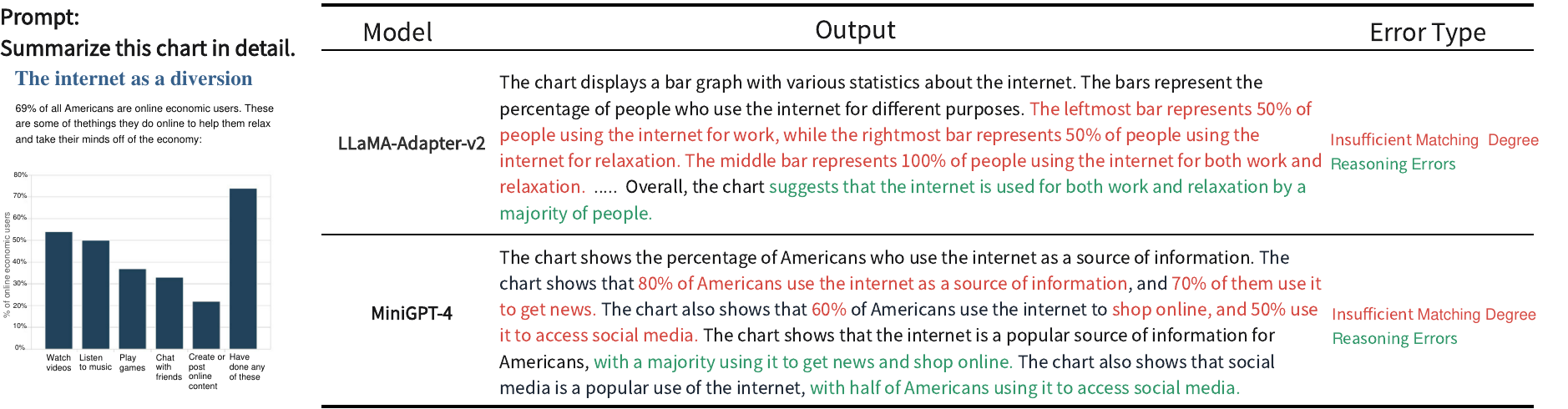} 
  \caption{Comparison with large visual-language models in chart summarization (LLaMA-Adapter-v2 \cite{gao2023llama}, MiniGPT-4 \cite{zhu2023minigpt}). There are two types of errors that occur during the generation process: Insufficient matching degree (inconsistency between the generated results and the chart content, such as content omission or fabricated content), and reasoning errors (inconsistency between the inferred meaning and the intended message of the chart).}
  \label{in}
\end{figure*}

However, large visual-language models still face challenges when generating chart-based textual descriptions. Two primary challenges include:
 (i) \textbf{Insufficient Matching Degree:} This refers to the degree to which the generated summaries align with a chart's numerical information. There are two main pitfalls here: incomplete descriptions and fabricated chart content. The former arises because the proportion of numbers and text in charts is relatively small, causing these details to be easily overlooked by the models, leading to content omissions. The latter, fabricated chart content, is an outcome of large visual-language models being influenced by unrelated information from their pre-training corpus, which causes them to produce content not relevant to the charts \cite{du2022survey}.
(ii) \textbf{Reasoning Errors:} Large models often underperform in chart reasoning tasks \cite{jiang2022understanding,bertolini2022testing}. Beyond numerical descriptions, models are expected to provide a holistic summary that captures the chart's intended meaning. However, some charts present vast amounts of numerical data and intricate curve patterns. This complexity poses challenges for models in deciphering the inherent meaning represented by the data, leading them to sometimes misinterpret the chart's intended message, causing reasoning errors \cite{wang2022locate}.

To address these limitations, we propose a new method named ChartThinker for training context-aware visual-language models, which leverages the chain of thought (CoT) and context retrieval for generating textual descriptions from charts. First, we pre-train the model using 595,955 chart-description pairs to enhance the matching degree, and subsequently fine-tune it using 8 million question-answer pairs, aiming to improve the model's accuracy and robustness in handling diverse charts and questions. Additionally, we introduce a Context-Enhanced CoT Generator module. This module fuses thought chains with context retrieval, incorporating increased logic and contextual information during the generation process, aiming to enhance the model's reasoning ability. Lastly, we employ a chart parsing module. This module combines the extracted underlying data with the prompt and feeds it as input to the CoT Generator, enhancing the model's \textbf{accuracy and matching degree} in interpreting chart data.

We conduct extensive empirical analysis to answer the following research questions (RQ):
\begin{itemize}[leftmargin=*]
  \item \textbf{RQ1:} Can answer reasoning benefit from introducing a chain of thought?
  \item  \textbf{RQ2:} How can context retrieval and chain of thought effectively interact with each other?
  \item \textbf{RQ3:} How does instruction fine-tuning improve the chart-summary matching degree?
\end{itemize}

Our main contributions are as follows:
\begin{itemize}[leftmargin=*]
  \item A large-scale chart dataset consisting of 595,955 chart-caption pairs and 8 million instruction-question pairs, covering a diverse range of visual styles and topics.
  \item An effective method for chart summarization, which leverages a context-enhanced CoT generator to integrate CoT with context retrieval.
  \item Extensive automatic and human evaluations that demonstrate the state-of-the-art (SOTA) performance of ChartThinker on various chart benchmarks.
  To facilitate further research, we release our dataset and codes at \href{https://github.com/Notonion/ChartThinker}{OpenChartThinker}.
\end{itemize}

\begin{figure*}[h!]
  \centering
  \includegraphics[width=0.91\textwidth]{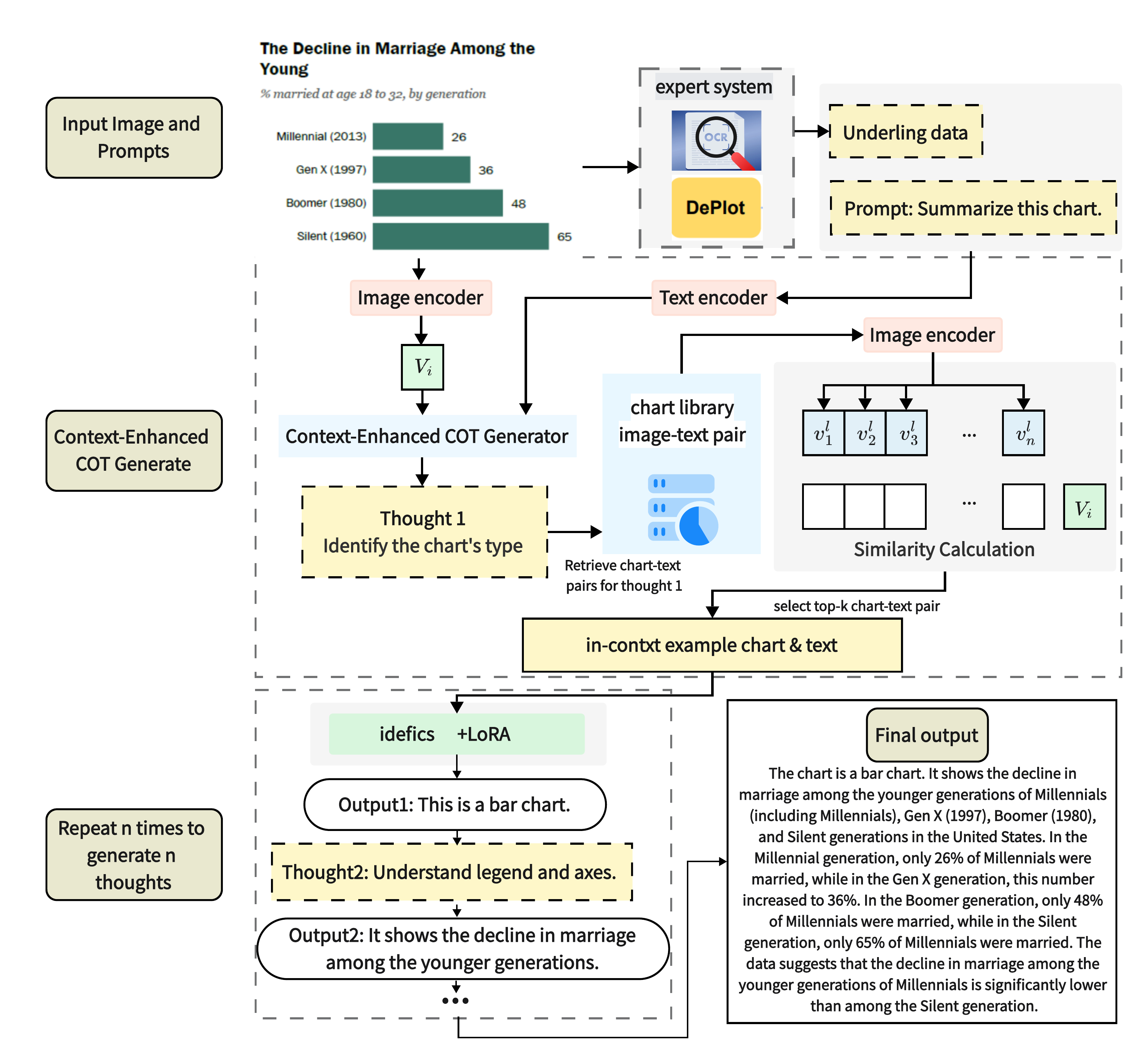} 
  \caption{Overview of ChartThinker. The encoded input chart and prompt are simultaneously fed into the Context-Enhanced CoT Generator. This module generates thought chains, and for each thought generated, the model retrieves the top-k image-text pairs from the chart library that best align with the thought, serving as contextual learning examples. Subsequently, the corresponding output for each thought is generated. Finally, all the outputs are consolidated to derive the final chart description.}
  \label{model}
\end{figure*}

\section{Related Work}
\textbf{Chart Summarization.}
Early methods for chart summarization relied on planning-based architectures \cite{reiter2007architecture}. The iGRAPH-Lite system \cite{ferres2007improving} employed template-based generation, aiding the visually impaired. Recent research by 
 \citet{chen2020logical} introduced a coarse-to-fine template-based generation method, and addressed some logical issues. While effective, the template-based generated text tends to be similar in nature, lacking individual differences and diversity.
To overcome these limitations, researchers began adopting architectures like LSTM \cite{hochreiter1997long} or transformers \cite{vaswani2017attention}. For instance, \citet{singh2020stl} and \citet{kantharaj2022chart} utilized ResNet to encode and introduced attention mechanisms, enhancing detail understanding. Additionally, some researchers \cite{obeid2020chart} have improved the factual accuracy of generated summaries by enhancing the embeddings in transformers.

To enhance the model's understanding of charts, \citet{liu2022matcha} made improvements based on Pix2Struct \cite{lee2023pix2struct} through pre-training on chart inverse rendering and mathematical reasoning tasks. 
In recent research, \citet{liu2022deplot} integrated the underlying information of charts with a large language model (LLM), improving the logical coherence and accuracy of chart summarization.


\textbf{Visual-Language Models.}
The field of LLMs has experienced significant advancements, exemplified by groundbreaking works such as ChatGPT \cite{brown2020language,openai2023gpt}, LLaMA \cite{touvron2023llama}, and Vicuna \cite{chiang2023vicuna}. More recently, multi-modal LLMs have garnered increasing attention. Flamingo \cite{alayrac2022flamingo} proposed a unified architecture with context-aware few-shot capabilities, and its open-source variant is OpenFlamingo \cite{awadalla2023openflamingo}. Blip2 \cite{li2023blip} bridges the modality gap between vision and language through a lightweight query transformer \cite{vaswani2017attention}. Mini-GPT4 \cite{zhu2023minigpt} builds on BLIP-2 to support longer responses and multi-turn dialogues better. LLaVA \cite{liu2023visual} employs a projection layer to align the frozen visual encoder CLIP \cite{radford2021learning} with the frozen LLM (Vicuna). LLaMA-Adapter \cite{zhang2023llama,gao2023llama} adapts LLaMA with additional adapter modules and multi-modal prompts. Ottor \cite{li2023otter} focuses on enhancing the model's ability to follow instructions through context examples.

\begin{table*}[h!]
\centering

\begin{tabular}{llcc}
\toprule
Dataset Name & Image Count & Q\&A Pair Count & Chart Types \\
\midrule
Autochart \cite{zhu2021autochart} & 6,003 & - & Scatter, Line, Bar \\
Linecap \cite{mahinpei2022linecap} & 3,528 & - & Line \\
DVQA \cite{kafle2018dvqa} & 300,000 & 3,480,000 & Scatter, Line, Bar \\
PlotQA \cite{methani2020plotqa} & 157,070 & 2,890,000 & Scatter, Line, Bar \\
Chart-to-text \cite{kantharaj2022chart} & 29,354 & - & Scatter, Line, Bar, Pie \\
FIGUREQA \cite{kahou2017figureqa} & 100,000 & 1,600,000 & Scatter, Line, Bar, Pie \\
\midrule
Chart-Sum-QA (Ours) & 595,955 & 8,170,000 & Scatter, Line, Bar,Pie \\
\bottomrule
\end{tabular}
\caption{Dataset Statistics.}
\label{dataset}
\end{table*}

\textbf{Prompt Engineering.}
Researchers have proposed various prompt engineering frameworks aimed at enhancing LLM reasoning, among which Prompt  Chain of Thought \cite{wei2022chain}, which guides the model's responses with intermediate reasoning examples, stands out as one of the most innovative and beneficial techniques. Its subsequent development, Chain-of-Thought-Self-Consistency \cite{wang2022self}, employs multiple reasoning paths, weighting them for optimized responses. Tree-of-Thoughts \cite{yao2023tree} showcases a tree-structured thought expansion, while Graph-of-Thoughts \cite{besta2023graph} progresses it into a directed acyclic graph, complete with self-loops. Algorithm-of-Thoughts \cite{sel2023algorithm} sets forth dynamic reasoning paths, mitigating redundancies. Skeleton-of-Thought \cite{ning2023skeleton} crafts a response blueprint, and Program-of-Thought \cite{chen2022program} articulates reasoning process into actionable programs.

\textbf{Our Position.} 
In summary, improving chart summarization with LLMs is under-explored. While current SOTA LLMs predominantly emphasize improving factual accuracy, logical coherence, or refining training architectures, they often overlook holistic integration. Our research bridges this gap by bolstering the accuracy and efficiency of vision-language LLMs specifically for chart summarization. We achieve this through the introduction of a novel dataset and a pioneering method that seamlessly blends Chain of Thought (CoT) with context retrieval strategies, all while maximizing context utilization.

\section{Methodology}
Given an input chart image $C$ and a prompt (question or instruction) $X$, we aim to generate an effective summary $\hat{S}$ that includes as much accurate information as possible, such as the chart's axes, data points, trends, and other relevant details. 

In this section, we begin by discussing the construction of the dataset (Sec. \ref{sec31}). Regarding the model architecture, we first utilize an image encoder and a text encoder to extract features from the input chart and prompt respectively (Sec. \ref{sec32}). Then, we introduce a chart parsing module that combines the obtained underlying data with the prompt to generate new text features (Sec. \ref{sec33}). Next, we design a Context-Enhanced CoT Generator module that integrates thought chains with context retrieval. By leveraging a small retrieval library, the model can access context examples related to the chart while constructing thought chains, injecting more logic and contextual information during the generation process (Sec. \ref{sec34}). 
Finally, all the generated thought chains are integrated with the LLM, Idefics \cite{idefics}, to produce the final output. 
The overall architecture of the model is shown in Figure \ref{model}.

\subsection{Dataset Construction}
\label{sec31}
We construct a dataset named \textbf{Chart-Sum-QA}, which includes comprehensive chart-summary pairs and question-answer pairs for instruction fine-tuning. Based on it, our model is pre-trained on 595,955 chart summary data points and is further fine-tuned using 8,170,000 instruction-question pairs (detailed in Table \ref{dataset}).

The process of constructing the dataset involves:

\textbf{(1) Data Collection:} We collect six different datasets containing images and their corresponding descriptions, consisting of 595,955 charts covering a broad range of topics and various chart types. These datasets are sourced from public image databases, research papers, or online image libraries. All datasets are covered under appropriate licenses (e.g., CC BY-NC-SA 4.0, MIT, GPL3.0).
\textbf{(2) Data Preprocessing:} For each dataset, we perform preprocessing to ensure the consistency and usability of the data \cite{chen2023data}. This includes resizing images, converting and standardizing formats, as well as cleaning and standardizing titles or descriptions. We ensure a precise alignment between images and their captions.
\textbf{(3) Generate Question-Answer Pairs:} To improve our model, we generate an additional 400,000 chart-question-answer pairs for instruction fine-tuning based on the summaries of Chart-to-text \cite{kantharaj2022chart}, Autochart \cite{zhu2021autochart}, and Linecap \cite{mahinpei2022linecap} datasets. The generated instruction fine-tuning dataset is merged with other QA datasets and filtered to obtain the final QA question-answer pairs for instruction fine-tuning, totaling 8,170,000 pairs, which include diverse questions about the charts. Since human annotations are costly, we generate questions automatically from human-written chart summaries using ChatGPT 4 and manually validate a subset of them for quality assurance. Through the QA dataset, the model can perform step-by-step learning more effectively in the contextual thought of chain, answering relevant questions more accurately.
\textbf{(4) Dataset Splitting:} We partition the dataset into 80\% training, 10\% validation, and 10\% testing.  The data is randomly and evenly distributed during the splitting process.

\begin{figure*}[h!]
  \centering
  \includegraphics[width=1\textwidth]{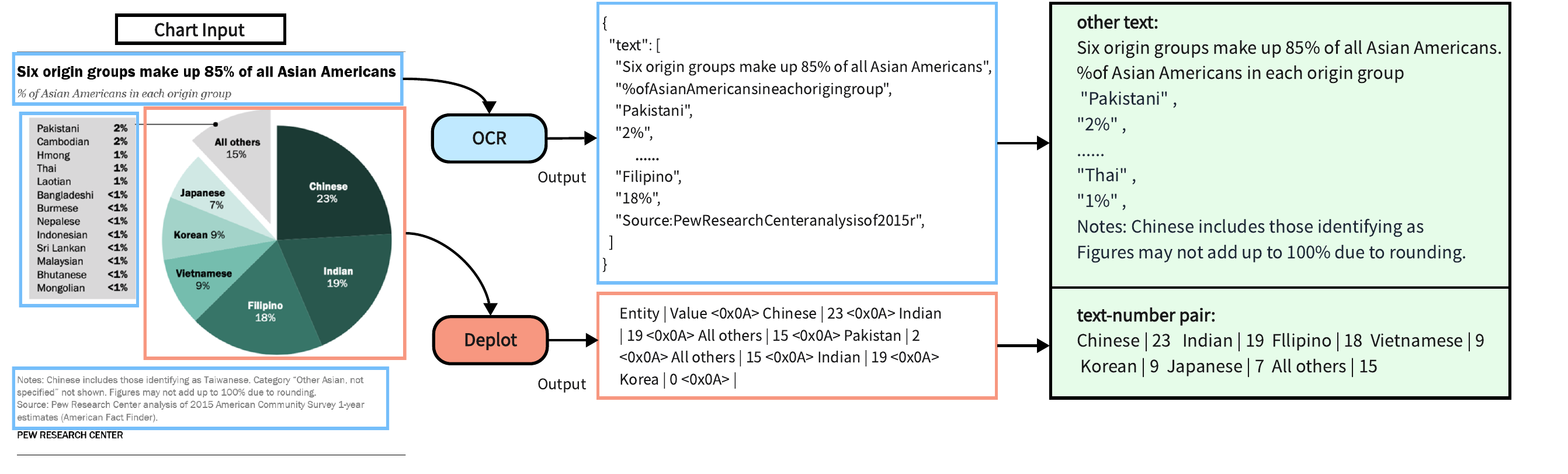} 
  \caption{The workflow of the chart analysis module. The input is the chart, the output of OCR is all of the textual and numerical information, and the output of Deplot is a table containing text and corresponding numerical data. The final integrated output is divided into two parts: text-number pair and other text.}
  \label{jiexi3}
\end{figure*}

\subsection{Image and Text Encoder}
\label{sec32}
Our chart image encoder is based on the encoder of CLIP \cite{radford2021learning}. The encoder takes an input chart and generates an encoded feature vector. To effectively encode chart images, the encoder identifies four components: (1) text elements (chart legends and axis labels), (2) data point elements (points, bars, lines representing specific values), (3) visual elements (chart type, colors), and (4) trend elements (patterns and trends of lines and scatter points in the chart). By recognizing and understanding these components, the chart image encoder generates a comprehensive encoded feature vector that captures the key information of the chart. The input chart image is processed by the encoder through operations such as convolution, pooling, and fully connected layers to transform it into a fixed-length encoded feature vector:

\begin{equation}\label{vencoder}
 V = f_{encoder}(C) ,
\end{equation}
 where $V$ represents the encoded feature vector, $f_{encoder}$ represents the chart image encoder function, and $C$ represents the input chart image.

We employ the LLaMA2 \cite{touvron2023llama} decoder to generate the output. The input is the prompt (question or instruction), and the output is the token sequence obtained by the encoder:

\begin{equation}\label{tencoder}
 K = f_{encoder}(X) ,
\end{equation}
 where $K$ represents the encoded token sequence, $f_{encoder}$ represents the text encoder function, and $X$ represents the input prompt.

 \subsection{Chart Parsing Module}
 \label{sec33}
The chart parsing module consists of two parts. The first part is the OCR module, which can extract textual and numerical content from charts. However, it lacks the ability to extract corresponding positions. The second part is the deplot module\cite{liu2022deplot}, which converts charts into tables. It is effective in handling the correspondence between numbers and text, but sometimes struggles with accurate numerical extraction and can be easily affected by irrelevant chart elements. We integrate these two modules to output the textual legends of the chart and text-number pairs containing key information. After extracting the necessary information, the outputs of the chart parsing module are combined with prompts and inputted into the context-enhanced CoT generator. An example of this module's operation is shown in Figure \ref{jiexi3}.

\subsection{Context-Enhanced CoT Generator}
\label{sec34}

To improve the quality and logical consistency of the generated results, we propose the Context-Enhanced CoT Generator module, which combines thought chains with context retrieval. Below are the detailed steps of this module:

\begin{itemize}[leftmargin=*]
\item {\textbf{Building a Small Retrieval Library}}: We created a small retrieval library containing 1,000 pairs of charts and text, where each pair includes basic information about the chart, image trends, x-y coordinates, etc. These context examples correspond to each stage of the COT, as detailed in Appendix \ref{app1}.

\item {\textbf{Similarity Computation}}: We employ cosine similarity to measure the similarity between the features of the input image and each context example in the retrieval library:

\begin{equation}\label{similarity1}
{similarity}_i = \frac{{T_i \cdot V}}{{|T_i| \cdot |V|}} ,
\end{equation}
where $T_i$ represents the feature vector of the $i$-{th} context example, $V$ represents the encoded feature vector of the input chart, ($\cdot $) denotes the dot product, ($||$) represents the norm of the vector.

\item {\textbf{Context Learning Weight Computation}}: Based on the results of the similarity computation, we assign context learning weights to each context example.Given the sensitivity of language models to the order of input prompts, images that are more similar in context learning (i.e., those ranked higher) are given more weight. That is, the weight of each image $x_i$ is inversely proportional to its relative position i. We define the weight as $1/i$.
That is, the weight of each image $x_i$ is inversely proportional to its relative position $i$. We define the weight as $\frac{1}{i}$.

The weighting function $W_{\ell}$ is as follows:

\begin{equation}\label{weight}
W_{\ell}\left(x_1, x_2, \ldots, x_N\right)=\sum_{i=1}^N \frac{1}{i} \cdot f_{\ell}\left(x_i\right),
\end{equation}

where the function $f_{\ell}\left(x_i\right)$ represents the influence of the $\ell$ image when generating he $i$ th language token.

\item {\textbf{Logic and Context Information Injection}}: During the generation process, we use the Idefics generation model. Given the image context, therefore, the conditional probability of the text $y$ can be expressed as:

\begin{equation}\label{ppp}
p(y \mid x)=\prod_{\ell=1}^L p\left(y_{\ell} \mid y_{<\ell}, W_{\ell}\left(x_1, x_2, \ldots, x_N\right)\right).
\end{equation}

The generation of each language token $y_l$ depends not only on the previous text tokens $y_{<l}$, but also on the weighted influence of the image context calculated by weighting function $W_l$. This method ensures the effective use of the image context information in the generation process.

\end{itemize}

In a nutshell, our objective is to leverage the integration of thought chains with context retrieval to enhance the Context-Enhanced CoT Generator module. This enhancement aims to facilitate the provision of comprehensive contextual information, and enhance the quality and logical consistency of the generated results. The process of generating a CoT chart summary is shown in Figure \ref{cot}. Some specific generation examples are shown in Appendix \ref{appa}.

\begin{figure}[htbp]
  \centering
  \includegraphics[width=0.5\textwidth ]{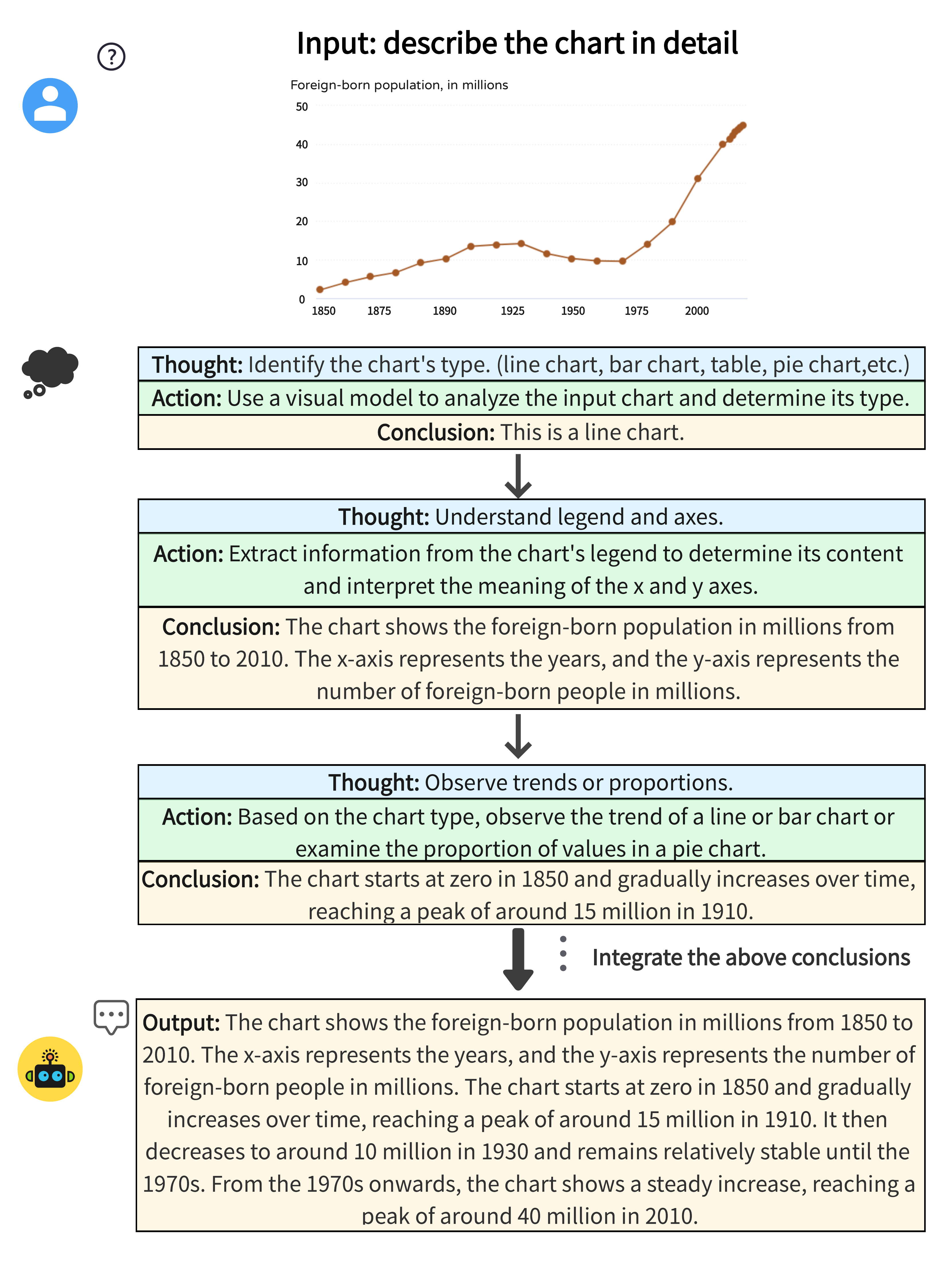} 
  \caption{The CoT Generation Process: For a given chart, the Context-Enhanced generator produces thoughts at each step. These thoughts help the model determine proper actions and generate conclusive statements. Finally, the conclusions from each step are integrated to yield the output answer.}
  \label{cot}
\end{figure}

\section{Experiment}

\subsection{Experimental Setup}

\begin{table*}
\centering
\small
\begin{tabular}{lccccccc} 
\toprule
& \textbf{BLEU} $\uparrow$ & \textbf{BLEURT}$\uparrow$ & \textbf{CIDEr}$\uparrow$ & \textbf{CS}$\uparrow$ & \textbf{PPL}$\downarrow$  &\textbf{$\bm{S_{\text{norm}}}$} $\uparrow$\\
\midrule
OCR-T5 \cite{raffel2020exploring} & 10.49 & -0.35  & 2.20 & \textbf{40.87\% }& 10.11 & 0.803\\
OCR-Chart2text \cite{obeid2020chart} & 7.2 & -0.56& 0.65 & 24.49\% & 12.11 & 0.338\\
OCR-Field-Infuse \cite{chen2020logical} & 0.19 & -1.01 & 0.26 & 10.12\% & 9.57 & 0.179\\
OCR-BART \cite{lewis2020bart} & 9.09 & -0.38 & 1.97  & 39.99 \%& 11.04 & 0.696\\
\midrule
OCR-ChartThinker (ours) & \textbf{11.81} & \textbf{-0.32} &\textbf{2.21} & 32.72\%& \textbf{9.23} & \textbf{0.948}\\
\bottomrule
\end{tabular}
\caption{Performance metrics compare our model with classic transformer-based pre-trained models, using both standard inputs (images and questions) and augmented inputs with OCR data from charts.}
\label{result1}
\end{table*}

\begin{table*}
\centering
\small
\begin{tabular}{lcccccc} 
\toprule
& \textbf{BLEU} $\uparrow$ & \textbf{BLEURT}$\uparrow$ & \textbf{CIDEr}$\uparrow$ & \textbf{CS}$\uparrow$ & \textbf{PPL}$\downarrow$ \\
\midrule
 LLaMA-Adapter-v2 \cite{gao2023llama} &1.07 & -0.83&0.36 &8.19\% &12.35 \\
 MiniGPT-4 \cite{zhu2023minigpt} &2.29 & -0.63 &0.62 & 11.77\%& 12.21\\
 mPLUG-Owl  \cite{ye2023mplug}&3.21 &-0.52 & 0.65 &16.9\% & 12.25\\
 LLaVA \cite{liu2023visual} &4.21 &-0.51 & 1.08& 19.15\%&12.16 \\
\midrule
ChartThinker (ours) & \textbf{5.82} & \textbf{-0.45} &\textbf{1.58} & \textbf{21.68\%}& \textbf{11.43 }\\
\bottomrule
\end{tabular}
\caption{Performance evaluation of our model versus large visual language models with an encoder-decoder framework, but not incorporating underlying data as input.}
\label{result2}
\end{table*}

\subsubsection{Baselines}
We compare our model against 8 baselines. (1) \textbf{T5} \cite{raffel2020exploring}: A unified seq2seq transformer model that achieved SOTA results on various text-to-text tasks. (2) \textbf{Chart2text} \cite{obeid2020chart}: An adapted transformer model specifically designed for chart-to-text translation. (3) \textbf{Field-Infusing Model} \cite{chen2020logical}: A transformer encoder-decoder model that generates target summaries and incorporates bounding box information for positional details. (4) \textbf{BART} \cite{lewis2020bart}: A seq2seq transformer model pre-trained with denoising objectives, which has shown effectiveness in text generation tasks. (5) \textbf{LLaMA-Adapter-v2} \cite{gao2023llama}: A parameter-efficient visual instruction model enable powerful multi-modal reasoning. (6) \textbf{MiniGPT-4} \cite{zhu2023minigpt}: It demonstrates powerful multi-modal capabilities similar to GPT-4 by aligning visual features with an advanced LLM. (7) \textbf{mPLUG-Owl} \cite{ye2023mplug}: A model that achieves powerful visual understanding, multi-turn dialogue capability, and knowledge reasoning. (8) \textbf{LLaVA} \cite{liu2023visual}: An end-to-end trained model that achieves a new SOTA accuracy on Science QA \cite{lu2022learn}.

\subsubsection{Automatic \& Human Evaluation}

 To verify the matching degree between generated text and charts, we employ five measures for automatic evaluation in our study. We use \textbf{BLEU} \cite{post2018call} for n-gram overlaps and \textbf{BLEURT} \cite{sellam2020bleurt}, specifically BLEURT-base-128, for fluency and content accuracy. \textbf{CIDEr} \cite{vedantam2015cider} evaluates the TFIDF weighted n-gram overlaps between the modelgenerated text and the reference, while \textbf{Content Selection (CS) score} \cite{wiseman2017challenges} measures how well the generated text aligns with the gold answer. Lastly, we use the GPT-2 Medium model \cite{radford2019language} to determine readability via \textbf{perplexity}. To holistically assess ChartThinker's performance, we newly calculated average normalized scores across five indicators, with the formula \textbf{$(\bm{S_{\text{norm}}}$}
= $\frac{S - S_{\text{worst}}}{S_{\text{best}} - S_{\text{worst}}} )$. 

To further evaluate the quality of the summaries, we conduct a manual assessment of 200 generated chart summaries. Annotators evaluated each summary based on two criteria: 
(i) \textbf{Matching Degree} (the data in the generated summary matches the chart with minimal data omission or fabrication)
(ii) \textbf{Reasoning Correctness} (the summary accurately infers the intended message or viewpoint from the chart).
Summaries were rated on a 1-5 scale, with 1 being the lowest and 5 being the highest, and presented randomly to avoid bias. The final score was the average given by the three evaluators.

\subsubsection{Fine-tuning Implementation} 
We apply the LoRA mechanism \cite{hu2021lora} into our model, setting the rank of the update matrices to 16, which reduced the size and number of trainable parameters. 
Specifically, the LoRA update matrices were applied to the modules ``$q_{proj}$'', ``$k_{proj}$'', and ``$v_{proj}$''. 
To control the magnitude of the LoRA updates, we set the scaling factor, Alpha, to 32 and implemented a dropout rate of 0.05 on the LoRA layers to mitigate over-fitting. The model was initialized with the weights of the pre-trained base model and was fine-tuned using the ``paged\_adamw\_8bit'' optimizer with a learning rate of 2e-4. 
To emulate larger batch sizes, a gradient accumulation step of 8 was used. During fine-tuning, evaluations were conducted every 20 steps.

\subsection{Main Results}

\subsubsection{Benchmark Results}
For a comprehensive evaluation, we compare our model to two types of methods, which cover different model architectures, processing of visual information, and input configurations.

\begin{table*}
\centering
\small
\begin{tabular}{lccccccc} 
\toprule
& \textbf{BLEU} $\uparrow$ & \textbf{BLEURT} $\uparrow$ & \textbf{CIDEr} $\uparrow$ & \textbf{CS} $\uparrow$ & \textbf{PPL} $\downarrow$ & \textbf{Human} $\uparrow$\\
\midrule

ChartThinker (ours) & \textbf{5.82} & \textbf{-0.45} &\textbf{1.58} & \textbf{21.68\%}& \textbf{11.43 }&\textbf{4.25}\\
\midrule
 
 No Chart Parsing Module & 4.87 & -0.50&1.12 & 18.29\%& 11.68&3.98\\
 No Context-Enhanced & 5.45 & -0.53&1.34 &20.10\% & 12.07&4.11\\
 No CoT & 5.10 &-0.57 &1.22 & 19.87\%& 11.97&3.92\\
 No Context-Enhanced CoT Generator& 4.59&-0.60 & 1.10& 19.55\%& 12.38&3.85\\
  No finetune on caption dataset & 4.36&-0.60 & 1.19& 19.03\%& 11.75&3.74\\
 No finetune on inrtruction dataset & 4.52&-0.63 & 1.27& 19.23\%&11.50 &4.03\\

\bottomrule
\end{tabular}
\caption{We carry out five ablation experiments: (1) omit the Chart Parsing Module, (2) exclude context retrieval and use only COT, (3) exclude CoT and use only context examples, (4) remove the entire Context-Enhanced CoT Generator, and (5) forgo the instruction fine-tuning dataset during model tuning. }
\label{aa}
\end{table*}

\textbf{Comparing with classic decoder-only models.}
Initially, we juxtapose our model against four classic pre-trained LLMs, as shown in Table \ref{result1}. Beyond the standard inputs of images and questions, we enrich the input data by integrating OCR-generated content from each chart. This augmented input feeds both the benchmark models and our own. Our experimental outcomes show that our model registers BLEU and CIDEr scores of 11.81 and 2.21, outstripping all baseline models. This performance underscores the superior matching degree of our generated text with the corresponding charts. Additionally, our model excels in BLEURT and PPL metrics, reflecting the enhanced readability and fluency of our generated summaries. To more intuitively and comprehensively evaluate the overall performance of ChartThinker compared to other classic models, we calculated the average normalized score \textbf{$S_{norm}$}
  for each model. This method normalizes all scores to a range between 0 and 1, allowing for direct and fair comparison between different metrics. The final results show that OCR-ChartThinker not only excels in individual metrics but also demonstrates the best overall performance. Overall, these findings emphasize the advantages of our model architecture over traditional transformer models.

\textbf{Comparing with encoder-decoder models.}
In a subsequent phase, we compare our model with other large visual language models using the encoder-decoder architecture, as seen in Table \ref{result2}. We train five baseline models on the pew dataset and evaluate them on its test set \cite{kantharaj2022chart}. The results highlight the superior performance of our model over other visual language models with similar frameworks. This observation directly addresses \textbf{RQ1}, emphasizing that \textit{embedding the reasoning chain into the model positively influences answer inference}. With respect to \textbf{RQ2}, our findings confirm that \textit{the synergy between context retrieval and the reasoning chain bolsters model performance}. Our model adeptly marries context retrieval with the reasoning chain, guaranteeing peak inference at every juncture. The model progressively generates the desired outcomes and makes timely adjustments, resulting in summaries that match more closely with the underlying charts. As a testament to this, our BLEU score witnesses an enhancement of 1.61 when juxtaposed against contemporary SOTA methods.

Our model undergoes instruction fine-tuning for each chart, allowing for a more accurate description of the actual values in the chart and further enhancing chart comprehension. Consequently, our model achieves the best PPL score. This provides evidence for \textbf{RQ3}, suggesting that \textit{using a directive dataset in fine-tuning enhances performance}.

\subsubsection{Human Evaluation Results}
Table \ref{human} showcases a manual evaluation of chart summarization. In this assessment, our ChartThinker model is benchmarked against eight advanced baseline models, which include those based on the classic transformer architecture as well as other prominent visual language
models. The assessment primarily centers on two key criteria: \textbf{matching degree} between the summaries and the charts, and \textbf{reasoning correctness}.
In terms of matching degree, the summaries generated by ChartThinker faithfully represent the data and information from the charts and show fewer errors. This indicates that our model significantly reduces data omissions and fabrications. Regarding reasoning correctness, the evaluators consistently favored our model. This demonstrates that ChartThinker excels in interpreting charts and making accurate inferences, capturing the core messages conveyed by the charts. Compared to other baseline models, ChartThinker holds a notable edge in this domain. More details are provided in Appendix \ref{app2}.

\begin{table}
\setlength{\tabcolsep}{4pt}
\footnotesize
\centering
\begin{tabular}{lcc} 
\toprule
 & \textbf{Matching} & \textbf{Reasoning} \\
& \textbf{Degree} $\uparrow$ & \textbf{Correctness} $\uparrow$  \\
\midrule
OCR-T5  & 3.96 & 4.11  \\
OCR-Chart2text  & 3.58 & 4.02  \\
OCR-Field-Infuse  & 2.13 & 3.29 \\
OCR-BART & 3.79 & 3.87   \\
 LLaMA-Adapter-v2 &2.63 & 2.97 \\
 MiniGPT-4 &2.92 & 2.85  \\
 mPLUG-Owl  &3.10 &3.26  \\
 LLaVA  &3.27 &3.34  \\
\midrule
ChartThinker (ours) & \textbf{4.32} & \textbf{4.27}\\
\bottomrule
\end{tabular}
\caption{Evaluation results compared to 8 baselines on chart-to-text testset \cite{kantharaj2022chart}. }
\label{human}
\end{table}

\subsection{Ablation Studies}
To further assess the impact of different parts on our model, we conduct ablation studies. The results are shown in Table \ref{aa}.

\textbf{The impact of ChartThinker component.}
On the component level, we find that removing
any major component (Chart Parsing Module and Context-Enhanced CoT Generator) would cause a performance drop. From Table \ref{aa}, we observe that: (1) Remove the Chart Parsing Module results in a significant decrease in the accuracy of describing the underlying data of the chart. (2) The removal of the context retrieval component from the generator significantly decreases the coherence and logicality of the generated text. This decline in language proficiency is attributed to the model's loss of contextual examples. (3) Similarly, the removal of the CoT component in the generator results in a decline in reasoning ability and a decrease in the comprehensiveness of the generated content. This is due to its lack of step-by-step generation and final integration process.

\textbf{Caption Dataset.} Excluding the chart description dataset leads to a decline in performance on the chart-to-text pew testset. Notably, BLEU decreases by 1.46, CIDEr by 0.39, and CS by 2.65\%. This indicates challenges in producing precise chart summaries without the dataset. Additionally, a reduced BLEURT and a higher PPL highlight the model's difficulty with unfamiliar chart layouts.

\textbf{Instruction Dataset.}
To further investigate the
impact of the instruction dataset, we excluded the dataset from our training. Without this fine-tuning, the model' degree of chart-summary matching weakens, occasionally generating unrelated content. This mismatching arises because the model relying on its pre-training parameters, fails to adapt to chart tasks. The drop in BLEU and CS scores reveals challenges in extracting pertinent details and reasoning accurately. 

\subsection{Case Study}

\subsubsection{Training Paradigms and Task-Specific Optimization}

Regarding the performance gap between multi-modal large models and text-only models, LMMs, such as our proposed \textbf{ChartThinker}, are pre-trained on unsupervised text-image pairs and are not fine-tuned for specific tasks, which contrasts with text-only models using OCR systems trained on task-specific supervised data. This accounts for the superior performance of OCR+LLM models on certain tasks \cite{jiao2024enhancing}, where the OCR system benefits from training data distribution that mirrors the test set. As an illustration, the OCR in ChartThinker, , Paddle OCR, does not specialize in chart summarization, unlike other OCR models referenced in Table \ref{result1}. This difference in training is why OCR+ChartThinker underperforms in terms of the CS indicator, as also observed in the literature \cite{liu2023hidden}.

\subsubsection{Limitations of Content Similarity (CS) as a Single Metric.}
In Table \ref{result1}, ChartThinker performs lower on the Content Similarity (CS) metric compared to certain OCR models. But evaluating generative models on the sole basis of CS is limiting, as CS only measures similarity to a standard summary and may not reflect the accuracy of model-generated content. For instance, ChartThinker might generate detailed and accurate summaries, but if the content extends beyond the standard answer, the CS score might suggest an inferior performance.
An example detailed below underscores this issue (Table \ref{tab:casestudies}): despite ChartThinker producing a detailed and accurate summary, its CS score (32.29\%) is lower than OCR-T5's (38.74\%), which is less descriptive. 

\begin{table}[h!]
\centering
\begin{tabular}{p{0.9\linewidth}}
\toprule
\textbf{Gold Summary:} The global temperature records over the past 100 years depict a temperature rise process from 14.5 °C in 1920 to 16.2 °C in 2020, with temperatures sharply rising by 1.2 °C between 1980 and 2020. 1998 and 2016 are the two hottest years on record. \\
\midrule
\textbf{Summary by ChartThinker:}This line chart provides a detailed display of the changes in global temperature over the past 100 years. The data shows that the average temperature increased from 14.5 °C in 1920 to 16.2 °C in 2020, showing a significant upward trend overall. Especially between 1980 and 2020, temperatures rose by nearly 1.2 °C, reflecting the acceleration of global warming. In addition, the chart also specifically indicates the two extremely high-temperature years of 1998 and 2016. \\
\midrule
\textbf{Summary by OCR-T5:}This chart shows the global temperature changes over the past 100 years. The average temperature has risen from 14.5 °C in 1920 to 16.2 °C in 2020, showing an overall upward trend. At the same time, the increase of 1.2 °C from 1980 to 2020 indicates that global warming is intensifying. \\
\bottomrule
\end{tabular}
\caption{Case Studies in Content Summarization}
\label{tab:casestudies}
\end{table}

\section{Conclusion}
We introduce Chart-Sum-QA, a comprehensive dataset tailored for chart summarization, and ChartThinker, a new method capable of training visual-language models with enhanced utilization of contextual information. 
ChartThinker integrates the chain of thought with context retrieval to enrich the summaries with rigorous logic. 
Both automated and human evaluations were conducted, demonstrating the superiority of our approach in chart summarization tasks over 8 SOTA methods and 7 evaluation metrics.
Further, through extensive ablations, we elucidate the effectiveness of each component and the helpfulness of our dataset. 
Our findings underscore the key role of CoT in reasoning and the criticality of context retrieval for semantic understanding.
We hope our released dataset, codes, and empirical results can shed light on more LLMs-based chart-summarition studies.

\newpage
\nocite{*}
\newpage
\section{Bibliographical References}\label{sec:reference}

\bibliographystyle{lrec-coling2024-natbib}
\bibliography{lrec-coling2024-example}
\newpage

\appendix

\section{Generated Summary Example}
\label{appa}
\begin{figure}[!h]
  \centering
  \includegraphics[width=0.5\textwidth]{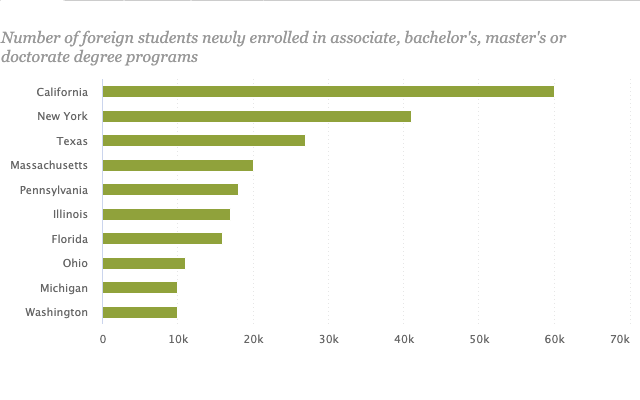} 
  \caption{This is a bar chart. The chart shows the number of foreign students newly enrolled in associate, bachelor's, master's, or doctorate degree programs in different states in the United States. California has the highest number of foreign students enrolled in a degree program, with 59,801 students enrolled. New York follows with 41,040 students, while Texas has 26,701 students. Massachusetts has 19,901 students, Pennsylvania has 17,801 students, Illinois has 16,801 students, Florida has 15,701 students, Ohio has 10,801 students, Michigan has 9,801 students, and Washington has 9,801 students. }
  \label{q1}
\end{figure}

\begin{figure}[!h]
  \centering
  \includegraphics[width=0.5\textwidth]{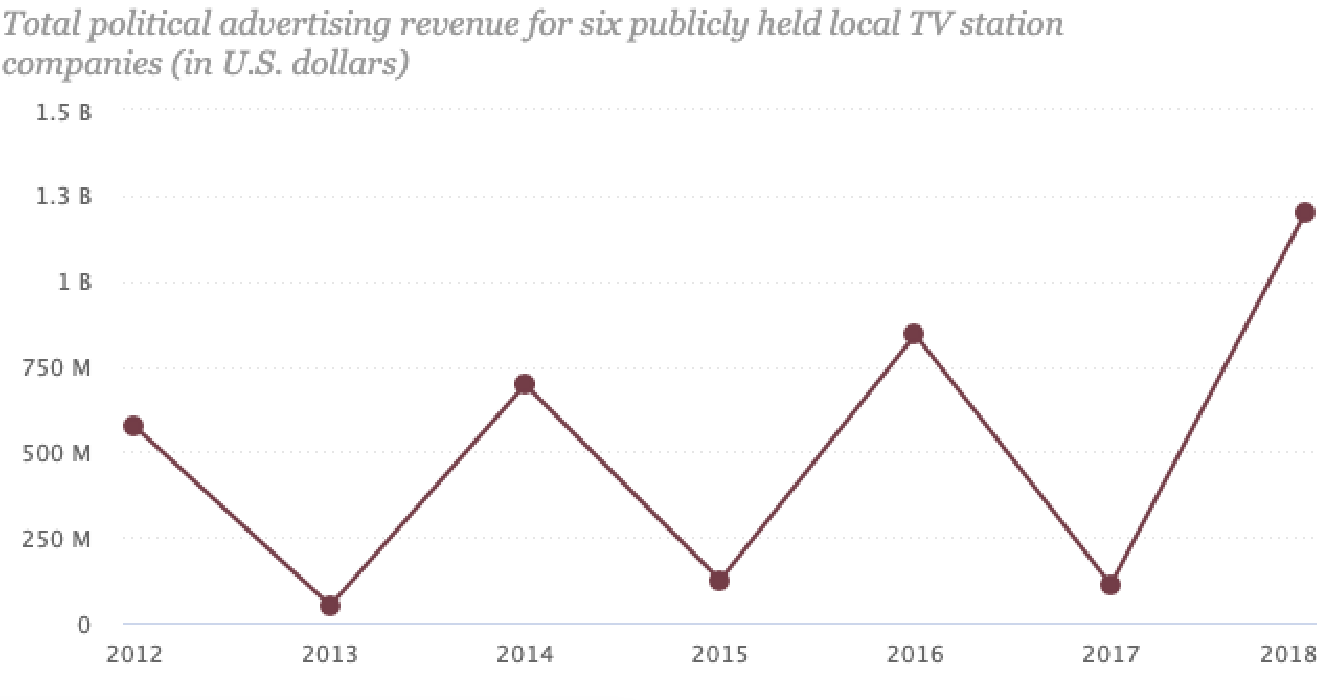} 
  \caption{This is a line chart. The chart shows the total political advertising revenue for six publicly held local TV station companies in the US from 2012 to 2018. The revenue decreased from 600 million in 2012 to 20 million in 2013, then increased to 700 million in 2014, decreased to 100 million in 2015, increased again to 800 million in 2016, decreased again to 100 million in 2017, and increased again to 1.3 billion in 2018. Therefore, the chart depicts a fluctuating trend of total political advertising revenue for six publicly held local TV station companies in the US during the six-year period, with an overall increasing trend from 2012 to 2018.}
  \label{q2}
\end{figure}

\begin{figure}[!h]
  \centering
  \includegraphics[width=0.5\textwidth]{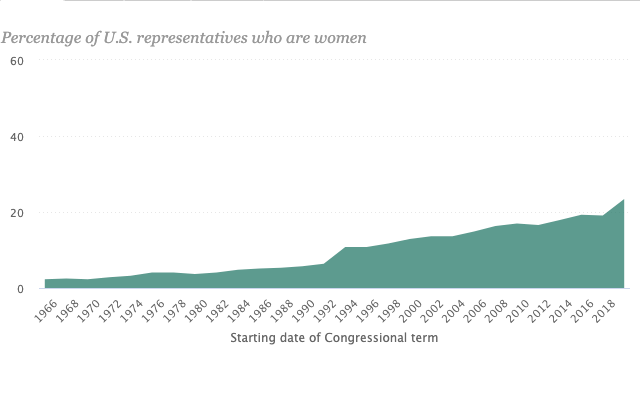} 
  \caption{This is a line chart. The chart shows the percentage of U.S. representatives who are women from 1966 to 2018.  The lowest percentage was 2.5\% in 1966 and there was an increase over the years. In 2018, the percentage was 24.1\%.  Overall, the chart shows that women represent a higher percentage than before. }
  \label{q3}
\end{figure}

\begin{figure}[!h]
  \centering
  \includegraphics[width=0.38\textwidth]{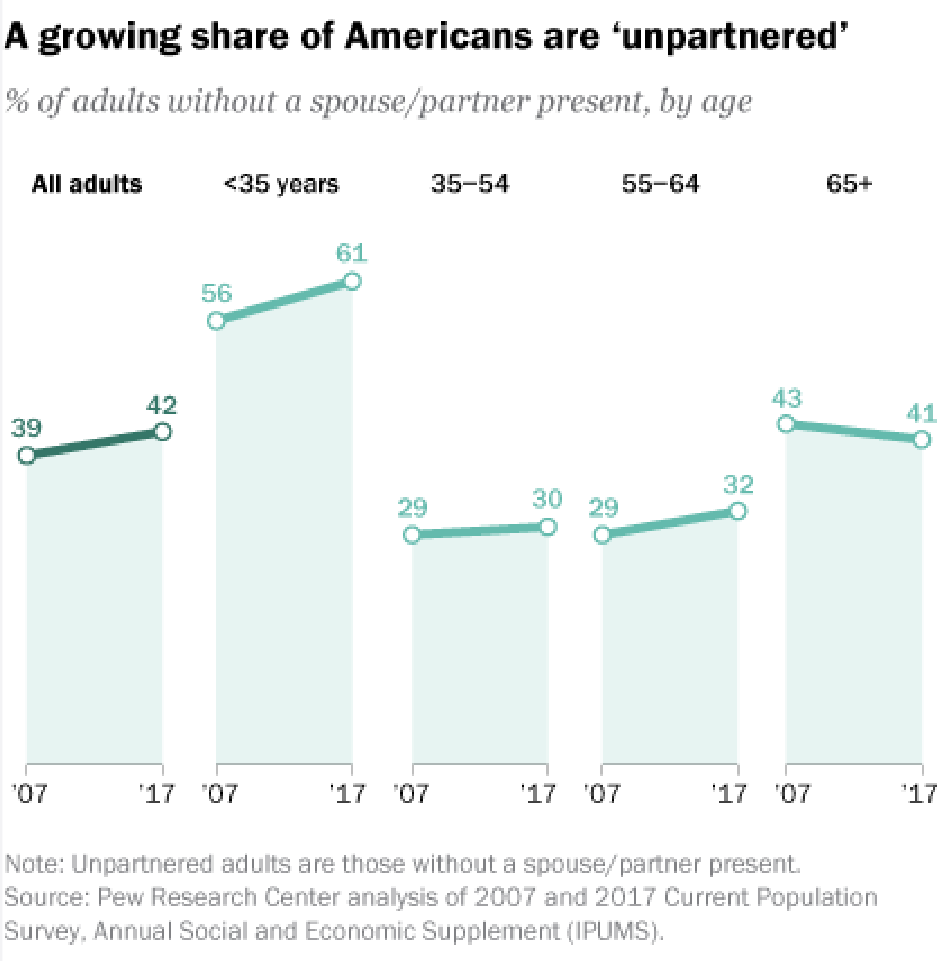} 
  \caption{This is a line chart. The chart shows the percentage of adults without a spouse/partner present in the United States who are unpartnered, broken down by age group from 2007 to 2017.  The percentage of all adults without a spouse/partner present increased slightly from 39\% in 2017 to 42\% in 2017. For the age under 35 years old, In 2007, 56\% of those without a spouse/partner present, compared to 61\% in 2017. For the age between 35 and 54 years old, In 2007, 29\% of those without a spouse/partner present, compared to 30\% in 2017. For the age between 55 and 64 years old, In 2007, 29\% of those without a spouse/partner present, compared to 32\% in 2017.  For the age over 65 years old, In 2007, 43\% of those without a spouse/partner present, compared to 41\% in 2017. Overall, the chart suggests that many Americans in the United States are unpartnered, with an increase in the percentage of those without a spouse/partner present.}
  \label{q5}
\end{figure}

\begin{figure}[!h]
  \centering
  \includegraphics[width=0.3\textwidth]{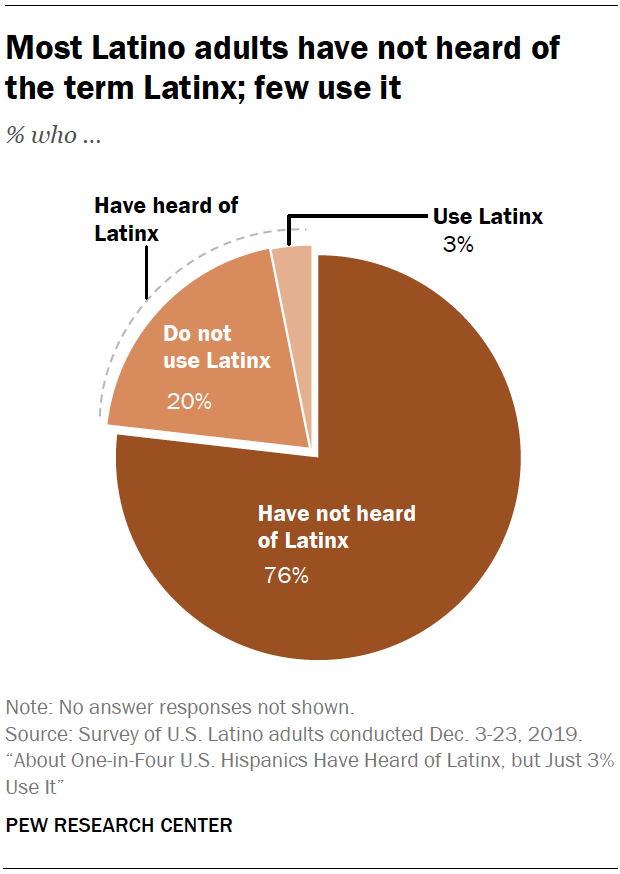} 
  \caption{This is a pie chart, according to a survey conducted by the Pew Research Center in December 2019,  most Latino adults have not heard of the term Latinx; few use it. It shows that 76\% of respondents said they have not heard of it, 3\% said they use Latinx, and 20\% said they have heard of it but did not use the term.}
  \label{q8}
\end{figure}

\begin{figure}[!h]
  \centering
  \includegraphics[width=0.4\textwidth]{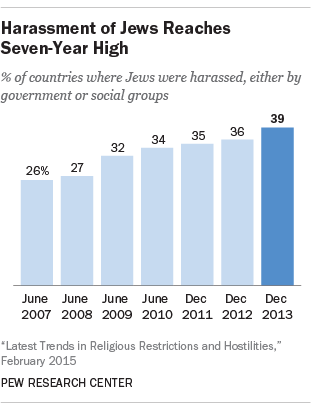} 
  \caption{This is a bar chart, the chart shows the harassment of Jews reached a seven-year high. The percentage has steadily increased over the years, with a peak of 39\% in December 2013 and a low of 26\% in June 2007. The number of harassment of Jews in 2008,2009, 2010, 2011, and 2012  is 27\%, 32\%,  34\% 35\%, and 36\% respectively. The data suggests that there has been a significant increase in harassment among Jews in the United States over the past few years.}
  \label{q10}
\end{figure}

In the following sections, we provide a series of summaries generated by ChartThinker, as illustrated in Figures \ref{q1} through \ref{q10}.

\begin{figure}[!h]
  \centering
  \includegraphics[width=0.35\textwidth]{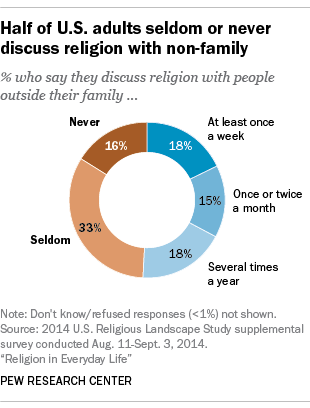} 
  \caption{This is a pie chart. The chart shows that half of U.S. adults seldom or never discuss religion with non-family.  According to the 2014 U.S. Religious Landscape Study conducted by the Pew Research Center, 18\% of U.S. adults believe that they discuss religion with people outside their family at least once a week, while 15\% believe it is at least once or twice a month, and 18\% believe it is several times a year. The remaining 33\% of respondents seldom discuss religion with non-family. And  16\% of them never talk about it.}
  \label{q7}
\end{figure}

\begin{figure}[!h]
  \centering
  \includegraphics[width=0.25\textwidth]{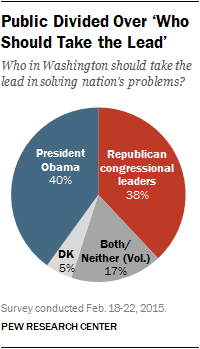} 
  \caption{This is a pie chart. According to a survey conducted by the Pew Research Center in February 2015, 40\% of respondents believed that president Obama should take the lead in solving the nation's problems, while 38\% believed that Republican congressional leaders should take the lead. Only 17\% of respondents believed that both/neither should take the lead, while 5\% believed that DK should take the lead.  }
  \label{q6}
\end{figure}

\begin{figure}[!h]
  \centering
  \includegraphics[width=0.4\textwidth]{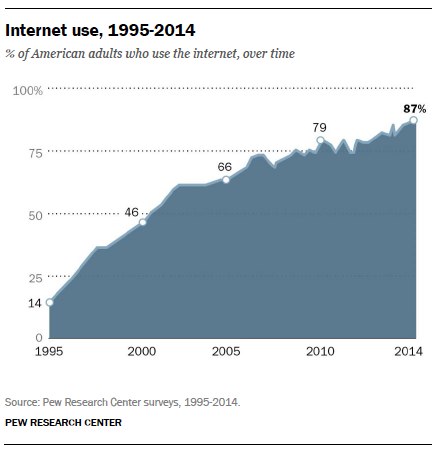} 
  \caption{This is a line chart. The chart shows the percentage of American adults who use the internet from 1995 to 2014. The number is 14\% in 1995, 46\% in 2000, 66\% in 2005, 79\% in 2010, 87\% in 2014. The chart shows that Internet usage has steadily increased over the years, reaching a peak of 87\% in 2014.}
  \label{q9}
\end{figure}

\begin{figure}[!h]
  \centering
  \includegraphics[width=0.5\textwidth]{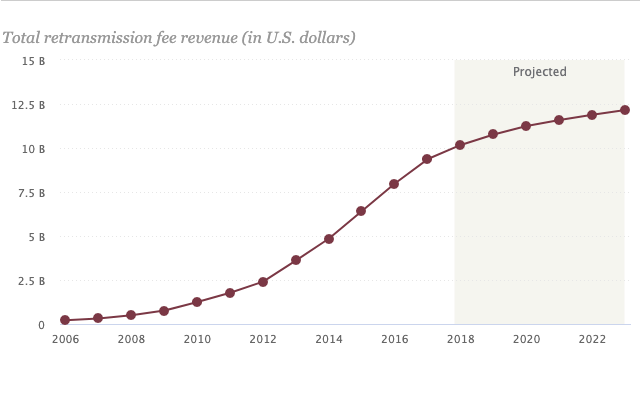} 
  \caption{This is a line chart. The chart displays the total retransmission fee revenue in U.S. dollars from 2006 to 2022. The revenue started at 0.1 billion dollars in 2006 and increased to 2.5 billion dollars in 2012.  The corresponding numbers are 5 in 2014, 8 in 2016, 10 in 2018, 11 in 2020 and 12.5 in 2022. Overall, the chart shows an increasing trend of total retransmission fee revenue in U.S. dollars from 2006 to 2022.}
  \label{q4}
\end{figure}

\section{Retrieval Library}
\label{app1}
\begin{figure*}[!h]
  \centering
  \includegraphics[width=0.95\textwidth]{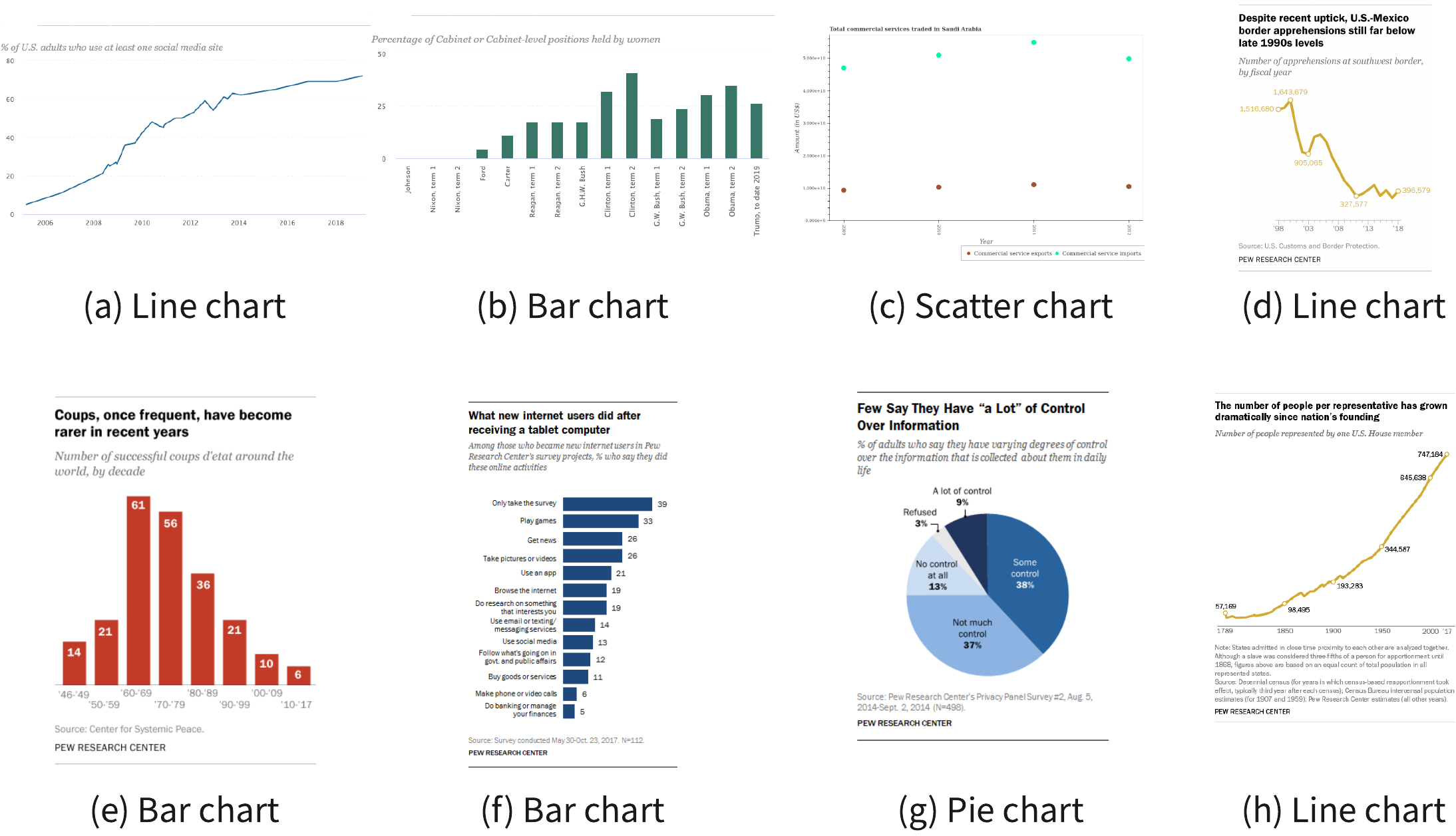} 
  \caption{Examples from the first stage of the Retrieval Library,  showcasing chart types for context retrieval.}
  \label{fulu3}
\end{figure*}

Our constructed context retrieval library is divided into four stages of examples. The first stage focuses on chart types. By analyzing a given chart, the model identifies and learns its type, with some examples presented in Figure \ref{fulu3}. 

\begin{figure*}[!h]
  \centering
  \includegraphics[width=1\textwidth]{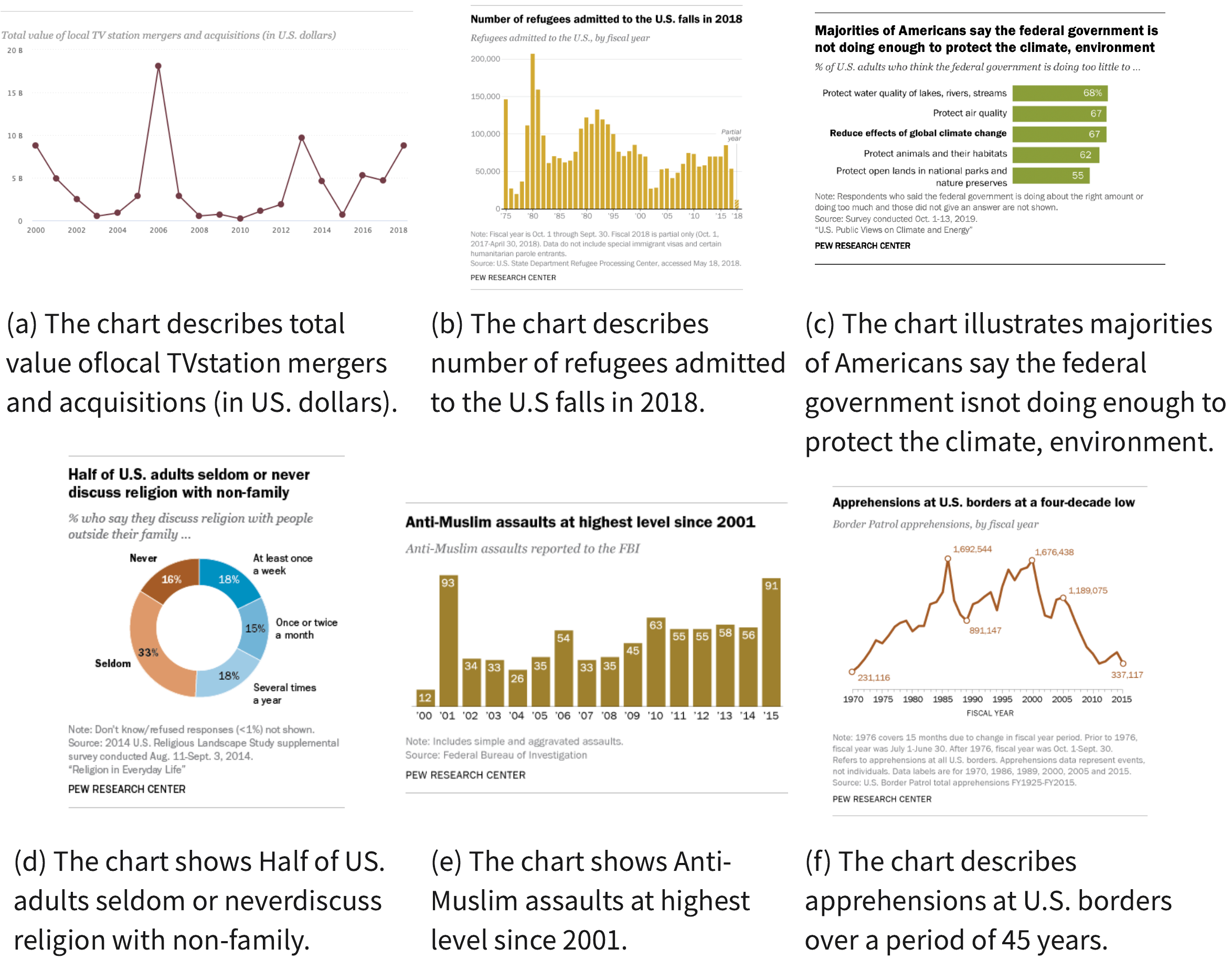} 
  \caption{Examples from the second stage of the Retrieval Library,  showcasing chart content overview in context retrieval.}
  \label{fulu4}
\end{figure*}

The second stage pertains to the overall caption of the chart, primarily derived from the chart's title. In this stage, the model learns from the context examples in the retrieval library to output a chart overview summary, as illustrated in Figure \ref{fulu4}.

\begin{figure*}[!h]
  \centering
  \includegraphics[width=0.99\textwidth]{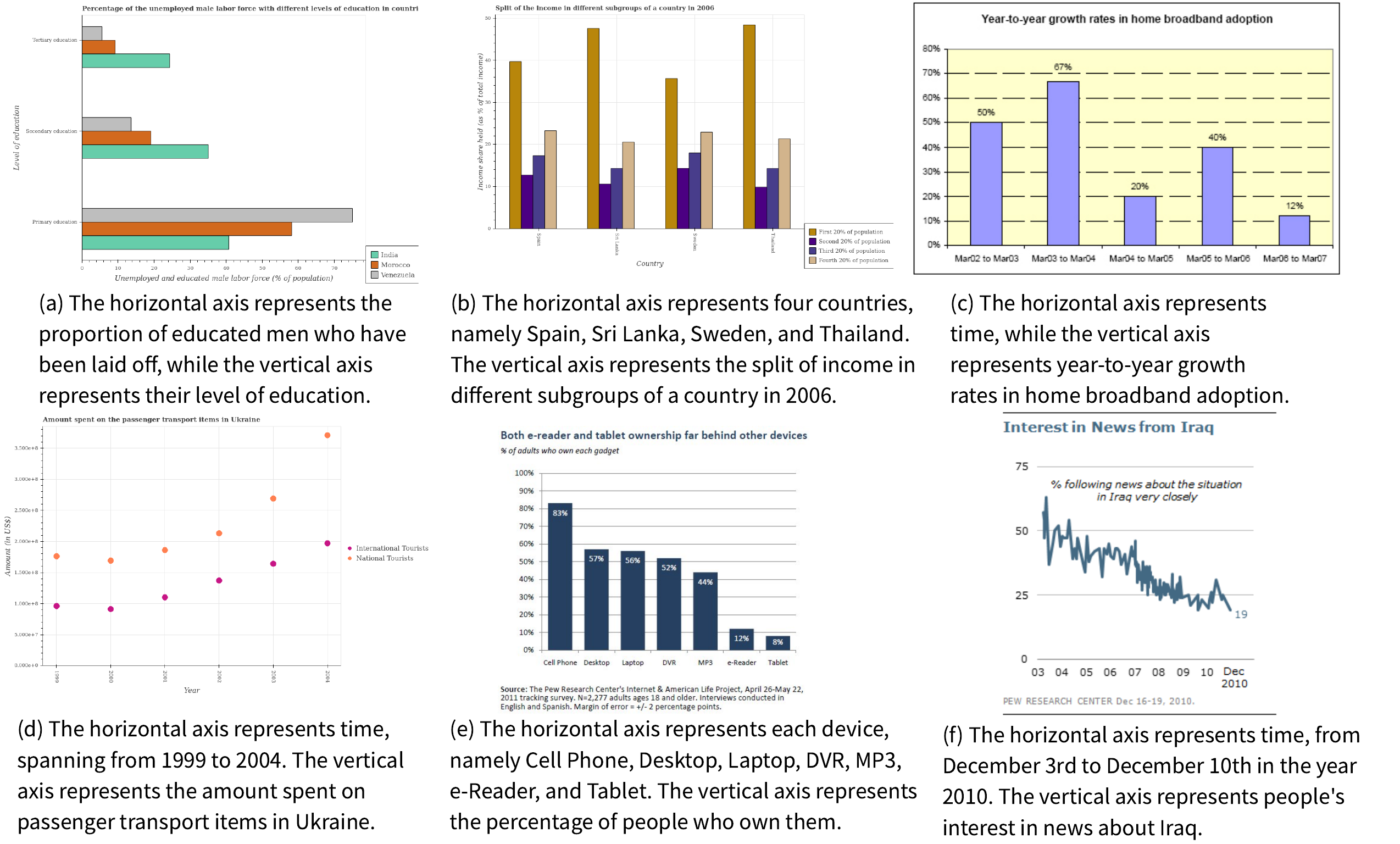} 
  \caption{The third stage in the Retrieval Library, which shows examples of Axes' Meaning.}
  \label{fulu5}
\end{figure*}

The third stage elucidates the meanings of both the horizontal and vertical axes of the chart. The primary objective here is to significantly deepen the model's understanding of the axes and the intricate relationships they share, as detailed in Figure \ref{fulu5}. 

\begin{figure*}[!h]
  \centering
  \includegraphics[width=0.97\textwidth]{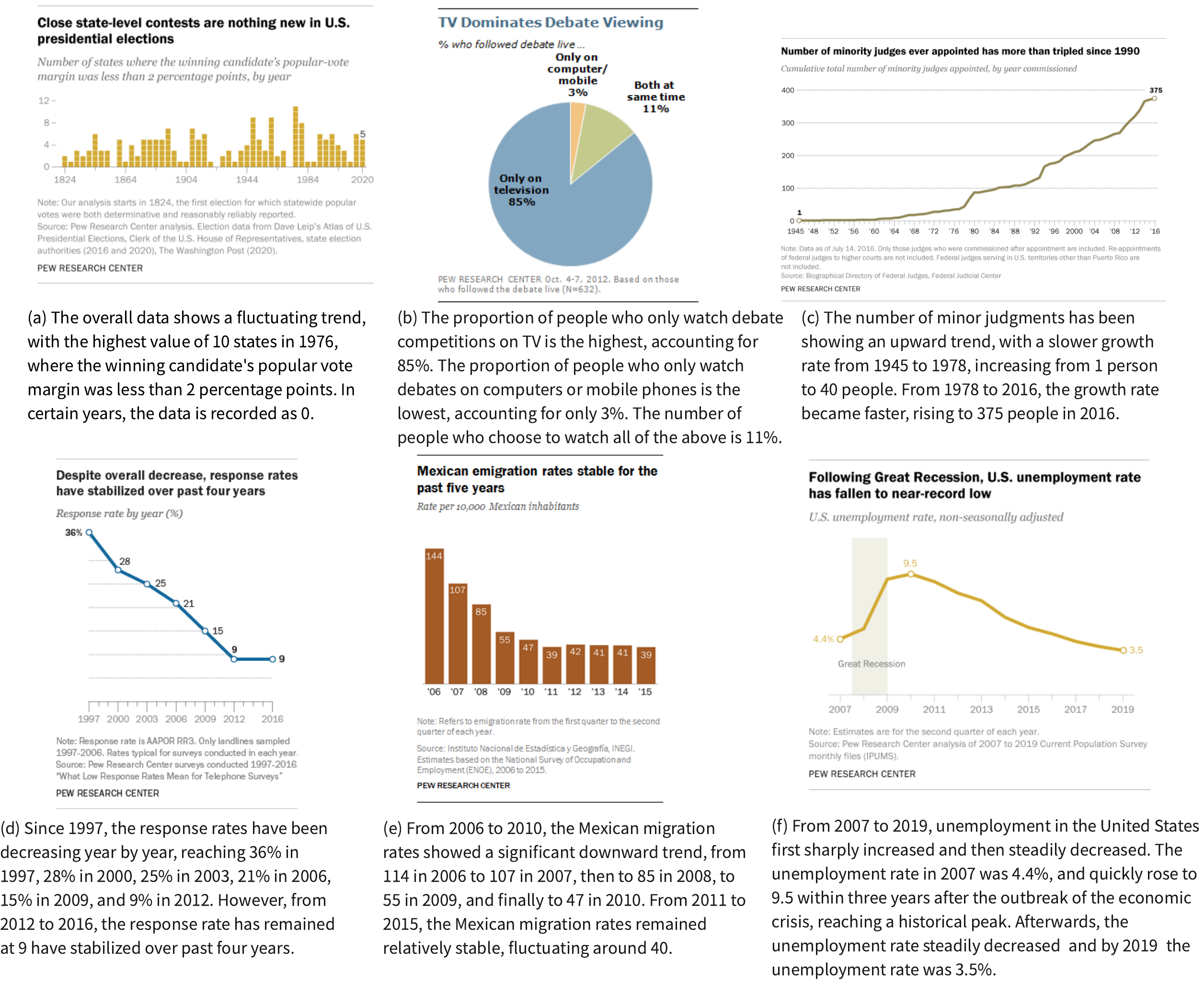} 
  \caption{The fourth stage in the Retrieval Library shows examples of Charts' numerical trend.}
  \label{fulu6}
\end{figure*}

The fourth stage centers on the trend of the chart data. The model learns to describe the chart's trend in natural language and generates a numerical trend description, as shown in Figure \ref{fulu6}.

To conclude, the context retrieval library comprises 1,000 charts, with each stage containing 250 charts and their associated textual descriptions. During training for each stage, the model searches within the 250 charts of the respective stage, selecting the top K most relevant charts as examples for context learning. Through this approach, the model becomes adept at describing charts in natural language, enhancing its chart comprehension and linguistic capabilities.


\section{Comparison of examples generated by various models}
\label{app2}

\begin{figure*}[!h]
  \centering
  \includegraphics[width=0.99\textwidth]{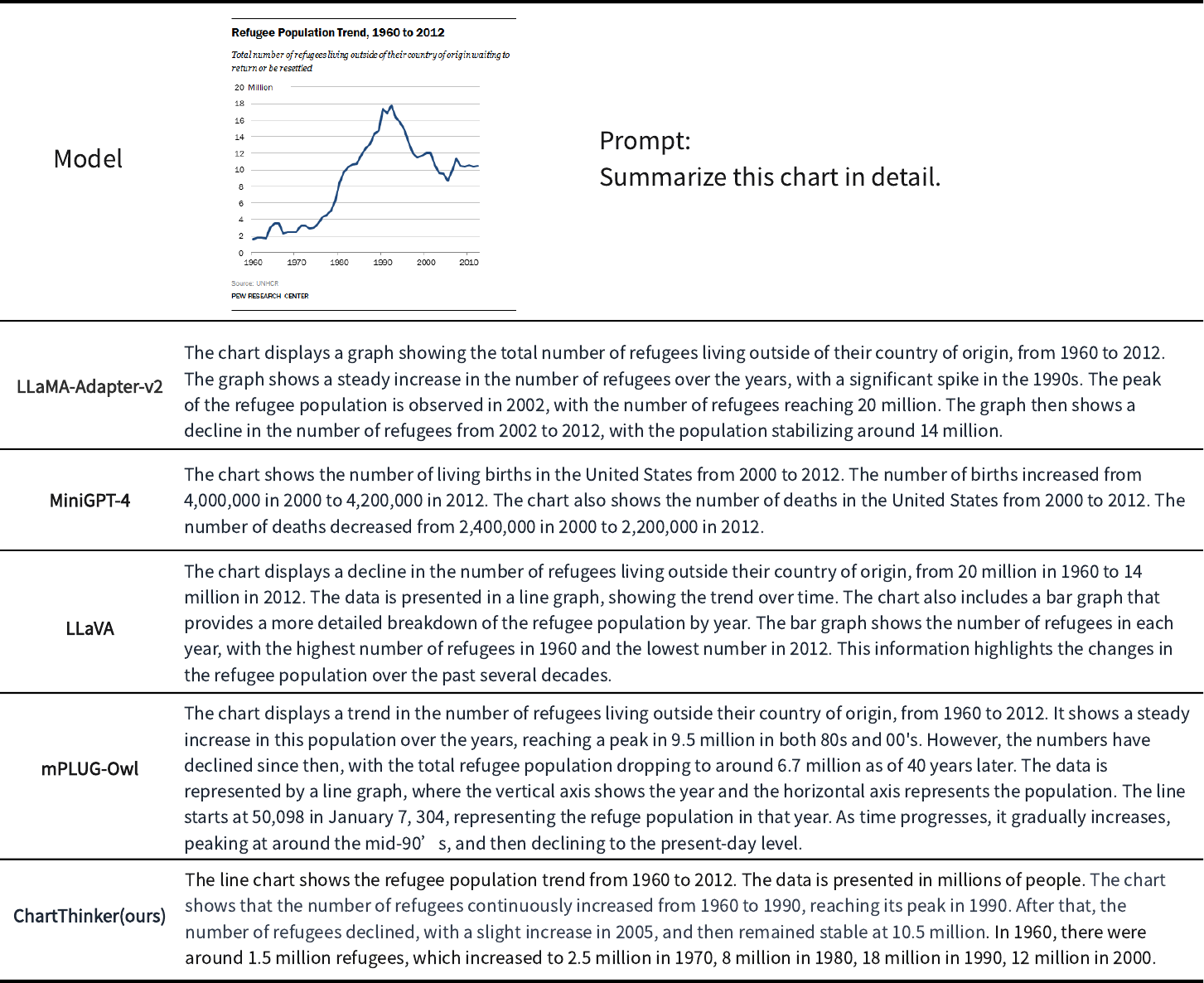} 
  \caption{Examples comparing the summaries generated by our model with those of other models.}
  \label{appi2}
\end{figure*}

In order to gain a more comprehensive understanding of our model's summaries, we undertake a meticulous manual assessment. The evaluation process involves evaluating a total of 200 summaries, including 40 summaries each from four baselines and our model. Figure \ref{appi2} showcases example summaries generated by the five models. The primary objective is to ensure summaries accurately and meaningfully represent the charts' essence and data.

To ensure an objective evaluation, we employ a set of annotators who are tasked with comparing each generated summary against its corresponding chart. The comparison is based on two pivotal criteria:
(i) \textbf{Matching Degree}: This criterion gauges the degree to which the data presented in the generated summary is in harmony with the chart.
(ii) \textbf{Reasoning Correctness}: Beyond just presenting data, it's imperative that the summary can accurately infer and convey the intended message or viewpoint that the chart aims to communicate.

\begin{figure*}[!h]
  \centering
  \includegraphics[width=0.9\textwidth]{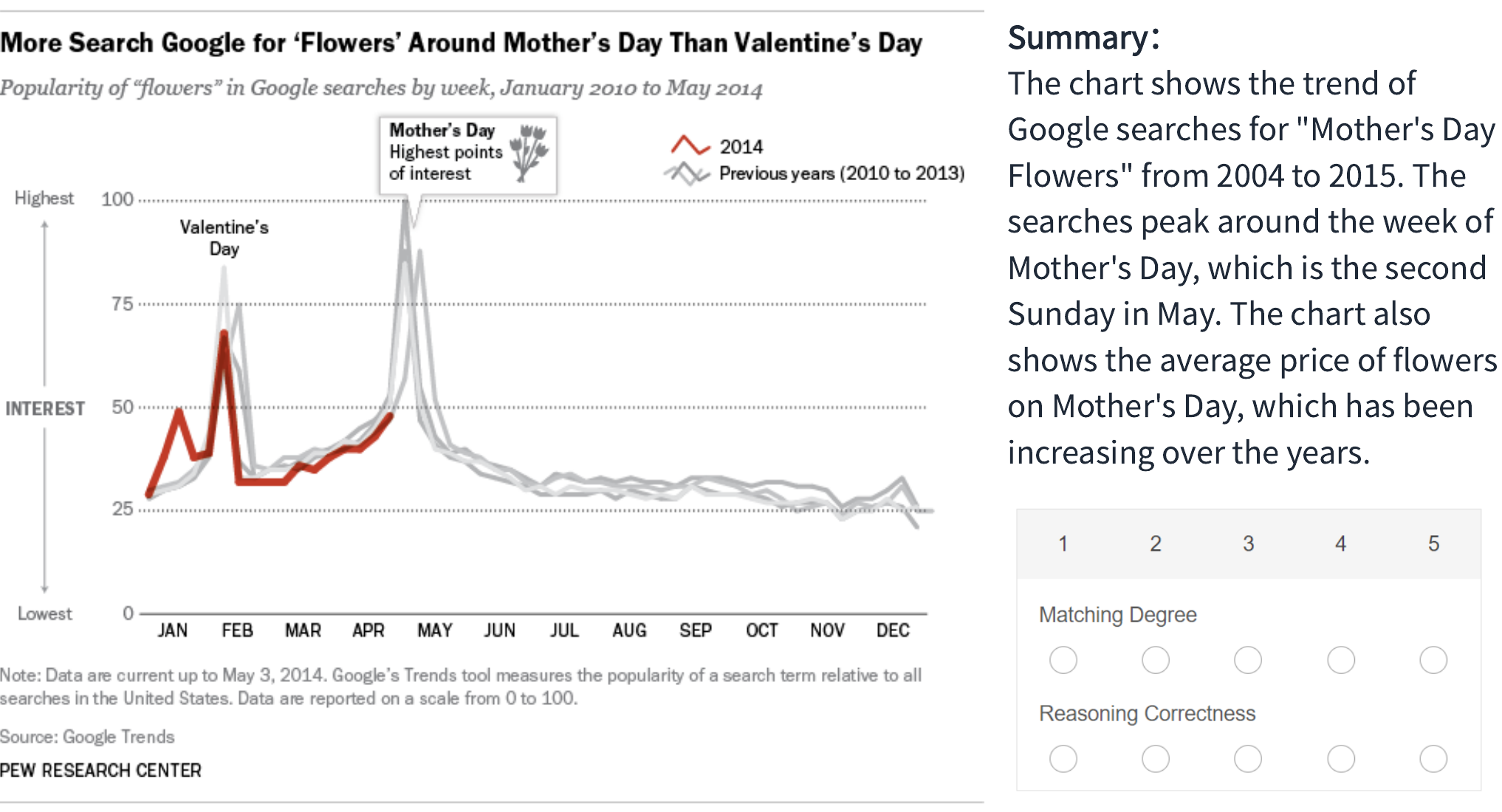} 
  \caption{Examples of human evaluation. For each chart, evaluators score the generated summary based on two criteria: matching degree and reasoning correctness, with scores ranging from 1 to 5.}
  \label{app3}
\end{figure*}

To uphold consistency and objectivity in the assessments, each summary is rated on a scale of 1 to 5, with 1 being the lowest and 5 being the highest. To further eliminate potential biases, the summaries are presented to the annotators in a random order. This strategy prevents evaluators from harboring preconceived notions or biases stemming from the presentation order. The complete evaluation procedure is illustrated in Figure \ref{app3}. After the evaluation, each summary's final score is determined by calculating the average of the scores given by three separate evaluators.

\bibliographystylelanguageresource{lrec-coling2024-natbib}
\bibliographylanguageresource{languageresource}

\end{document}